\documentclass[letterpaper, 10pt, journal, twoside]{IEEEtran}

\IEEEoverridecommandlockouts

\usepackage[dvipsnames,table]{xcolor}

\usepackage[utf8]{inputenc}
\usepackage[T1]{fontenc}
\usepackage[english]{babel}

\usepackage{comment}
\usepackage{xspace}
\usepackage{graphicx}
\usepackage{overpic}
\usepackage{svg}
\usepackage{bm}
\usepackage[hang,flushmargin]{footmisc}
\usepackage{booktabs}
\usepackage{tabularx}
\usepackage[nolist]{acronym}
\usepackage{soul}
\usepackage{subfig}
\usepackage{placeins}

\usepackage{tabularx}
\usepackage{multirow}
\usepackage{booktabs}
\usepackage{colortbl}
\usepackage{float}
\usepackage{placeins}

\usepackage{array}
\usepackage{nicematrix}
\newcommand{\PreserveBackslash}[1]{\let\temp=\\#1\let\\=\temp}
\newcolumntype{s}{>{\hsize=0.1}X}
\newcolumntype{C}[1]{>{\centering}p{#1}}
\newcolumntype{R}[1]{>{\raggedright}p{#1}}
\newcolumntype{L}[1]{>{\raggedright}p{#1}}
\newcolumntype{P}[1]{>{\centering\arraybackslash}p{#1}}  %

\usepackage{etoolbox}
\makeatletter
\patchcmd{\@makecaption}
{\scshape}
{}
{}
{}
\newcommand{\onetagright}{\tagsleft@false}
\makeatother

\usepackage{amsmath}
\usepackage{amssymb}
\usepackage{amsthm}
\usepackage{siunitx}

\newtheoremstyle{main}
{1em}                                                %
{1em}                                                %
{\itshape}                                        %
{0pt}                                                %
{\scshape}                                           %
{\\*}                                                %
{2pt}                                                %
{\thmname{#1}\thmnumber{ #2}: \thmnote{\itshape #3}} %

\usepackage{environ}

\usepackage[linesnumbered,ruled,vlined]{algorithm2e}
\newcommand{\removelatexerror}{\let\@latex@error\@gobble}

\usepackage[
	activate   = {true},
	protrusion = false,
	expansion  = true,
	kerning    = true,
	spacing    = true,
	tracking   = false,
	auto       = true,
	selected   = true,
	factor     = 2000,
	stretch    = 50,
	shrink     = 20,
]{microtype}
\usepackage{accsupp}

\usepackage{csquotes}
\usepackage[
	maxbibnames=99,
	maxcitenames=2,
	natbib=true,
	style=numeric-comp,
	backend=biber,
	sorting=none,
	giveninits=true,
	url=false,
	doi=false,
	eprint=false,
	isbn=false,
]{biblatex}

\makeatletter
\let\NAT@parse\undefined
\makeatother
\usepackage[pdfa,colorlinks,bookmarksopen,bookmarksnumbered,allcolors=black]{hyperref}

\usepackage[nameinlink,capitalise]{cleveref}
\crefname{line}{Line}{Lines}
\crefname{figure}{Fig.}{Figs.}
\Crefname{figure}{Fig.}{Figs.}
\crefname{equation}{Eq.}{Eqs.}
\Crefname{equation}{Eq.}{Eqs.}
\crefname{section}{Sec.}{Secs.}
\Crefname{section}{Sec.}{Secs.}
\crefname{definition}{Def.}{Defs.}
\Crefname{definition}{Def.}{Defs.}
\crefname{algorithm}{Alg.}{Algs.}
\Crefname{algorithm}{Alg.}{Algs.}
\crefname{assumption}{Asm.}{Asms.}
\Crefname{assumption}{Asm.}{Asms.}
\crefname{subassumption}{Asm.}{Asms.}
\Crefname{subassumption}{Asm.}{Asms.}
\Crefname{problem}{Problem}{Problems}
\crefname{problem}{Problem}{Problems}

\usepackage{flushend}

\usepackage[inline]{enumitem}
\usepackage{mathtools}

\newcommand{\eg}{\emph{e.g.},\xspace}
\newcommand{\ie}{i.e.,\xspace}

\newcommand{\squeezeWords}{\looseness=-1}
\newcommand{\squeezeLine}{\enlargethispage*{\baselineskip}\pagebreak}

\DeclareMathOperator*{\argmin}{arg\,min}

\DeclareMathAlphabet{\pazocal}{OMS}{zplm}{m}{n}
\newcommand{\unif}{\pazocal{U}}
\newcommand{\newAlgoLine}[1]{{\color{BrickRed} #1}}

\graphicspath{ {./images/} }

\makeatletter
\newcommand\notsotiny{\@setfontsize\notsotiny{6.2}{7}}
\makeatother

\definecolor{lightergray}{gray}{0.925}

\begin{acronym}
    \acro{VAMP}{Vector-Accelerated Motion Planning}
    \acro{OMPL}{Open Motion Planning Library}
    \acro{ASAO}[a.s.a.o.]{almost-surely asymptotically optimal}
    \acro{asao}{almost-surely asymptotically optimal}
    \acro{RRT}{Rapidly-exploring Random Trees}
    \acro{G-RRT*}{Greedy RRT*}
    \acro{SIMD}{single instruction/multiple data}
    \acro{BIT*}{Batch Informed Trees}
    \acro{B-RRT*}{Bidirectional RRT*}
    \acro{AIT*}{Adaptively Informed Trees}
    \acro{EIT*}{Effort Informed Trees}
    \acro{PRM}{Probabilistic Roadmaps}
    \acro{ABIT*}{Advanced BIT*}
    \acro{GBIT*}{Greedy BIT*}
    \acro{FIT*}{Flexible Informed Trees}
    \acro{MBM}{MotionBenchMaker}
    \acro{RGG}{random geometric graph}
    \acro{FCIT*}{Fully Connected Informed Trees}
    \acro{EST}{Expansive Space Tree}
    \acro{TB-RRT}{Time-based RRT}
    \acro{ST-RRT*}{Space-Time RRT*}
    \acro{SI-RRT}{Safe-Interval RRT}
    \acro{WHCA*}{Windowed Hierarchical Cooperative A*}
    \acro{SIPP}{Safe Interval Path Planning}
    
    \acro{AOX}{AO-\emph{x}}
    \acro{AORRTC}{Asymptotically Optimal RRT-Connect}
    
    \acro{DoF}{degree-of-freedom}
    \acro{RGG}{random geometric graph}
\end{acronym}

\title{AORRTC: Almost-Surely Asymptotically Optimal Planning with RRT-Connect
}
\author{Tyler S.\ Wilson$^{1}$, Wil Thomason$^{2}$, Zachary Kingston$^{3}$, and Jonathan D.\ Gammell$^{1}$
\thanks{Manuscript received: May, 15, 2025; Revised August, 10, 2025; Accepted September, 11, 2025.}%
\thanks{This paper was recommended for publication by Editor Júlia Borràs Sol upon evaluation of the Associate Editor and Reviewers' comments.}
\thanks{This work was supported by the Natural Sciences and Engineering Research Council of Canada (NSERC) [RGPIN-2024-06637].}%
\thanks{$^{1}$Estimation, Search, and Planning (ESP) Research Group, Queen's University, Kingston ON, Canada. \texttt{\{18tsw1,gammell\}@queensu.ca}}%
\thanks{$^{2}$\texttt{wil.thomason@gmail.com}}%
\thanks{$^{3}$Department of Computer Science, Purdue University, West Lafayette IN, USA \texttt{zkingston@purdue.edu}}%
\thanks{Digital Object Identifier (DOI): see top of this page.}
}

\begin{document}
    \maketitle

    \begin{abstract}
    
    Finding high-quality solutions quickly is an important objective in motion planning.
    This is especially true for high-\acl{DoF} robots.
    Satisficing planners have traditionally found feasible solutions quickly but provide no guarantees on their optimality, while \ac{ASAO} planners have probabilistic guarantees on their convergence towards an optimal solution but are more computationally expensive.
    \squeezeWords

    This paper uses the \acl{AOX} meta-algorithm to extend the satisficing RRT-Connect planner to optimal planning.
    The resulting \ac{AORRTC} finds initial solutions in similar times as RRT-Connect and uses additional planning time to converge towards the optimal solution in an anytime manner.
    It is proven to be probabilistically complete and \ac{ASAO}
    \squeezeWords

    \ac{AORRTC} was tested with the Panda (7 DoF) and Fetch (8 DoF) robotic arms on the \acl{MBM} dataset.
    These experiments show that \ac{AORRTC} finds initial solutions as fast as RRT-Connect and faster than the tested state-of-the-art \ac{ASAO} algorithms while converging to better solutions faster.
    \ac{AORRTC} finds solutions to difficult high-DoF planning problems in \emph{milliseconds} where the other \ac{ASAO} planners could not consistently find solutions in seconds.
    This performance was demonstrated both with and without \ac{SIMD} acceleration.

    \end{abstract}

    \acresetall

    \section{Introduction} \label{sec:intro}

    \IEEEPARstart{M}{otion} planning seeks to quickly find high-quality solutions to a given problem, especially when planning for high \ac{DoF} robots or in real time.
    Motion planning algorithms search a discrete approximation of the robot's continuous \emph{configuration space} (\ie search space).
    \squeezeWords
    
    Motion planning algorithms approximate the search space in different ways.
    Graph-based planners, such as Dijkstra's algorithm~\cite{dijkstra} and A*~\cite{astar}, require \emph{a~priori} discretization of the search space.
    High-resolution approximations usually contain high quality solutions but are computationally expensive to search, while low-resolution approximations are cheaper to search but may only contain low quality solutions, or no solution at all.
    Trajectory optimization methods, such as CHOMP~\cite{chomp} and TrajOpt~\cite{trajopt}, are less dependent on their approximation but only provide local guarantees and may not find a solution when getting stuck in local minima on difficult planning problems~\cite{trajopt}.

     \begin{figure}[t]

    \subfloat [Computed Motions (Fetch)] {%
        \includegraphics[width=.33\linewidth]{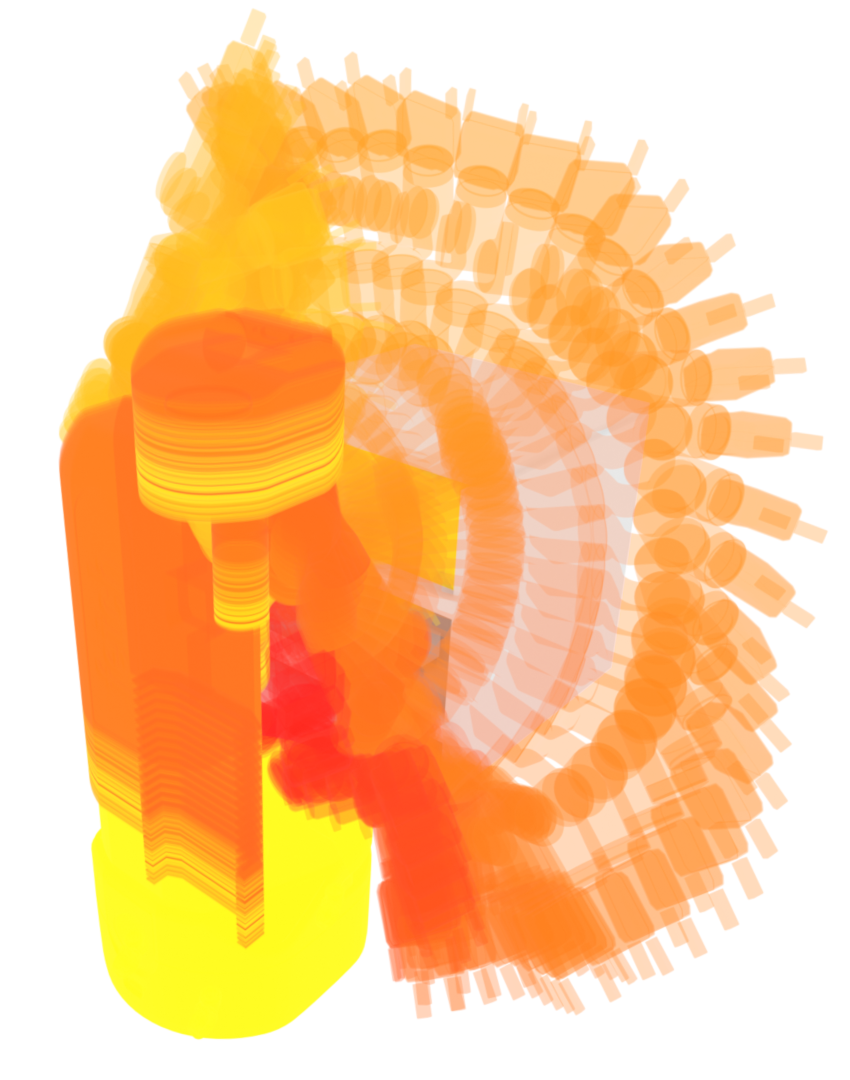} \label{fig:cage_initial_all:motions}
        \includegraphics[width=.33\linewidth]{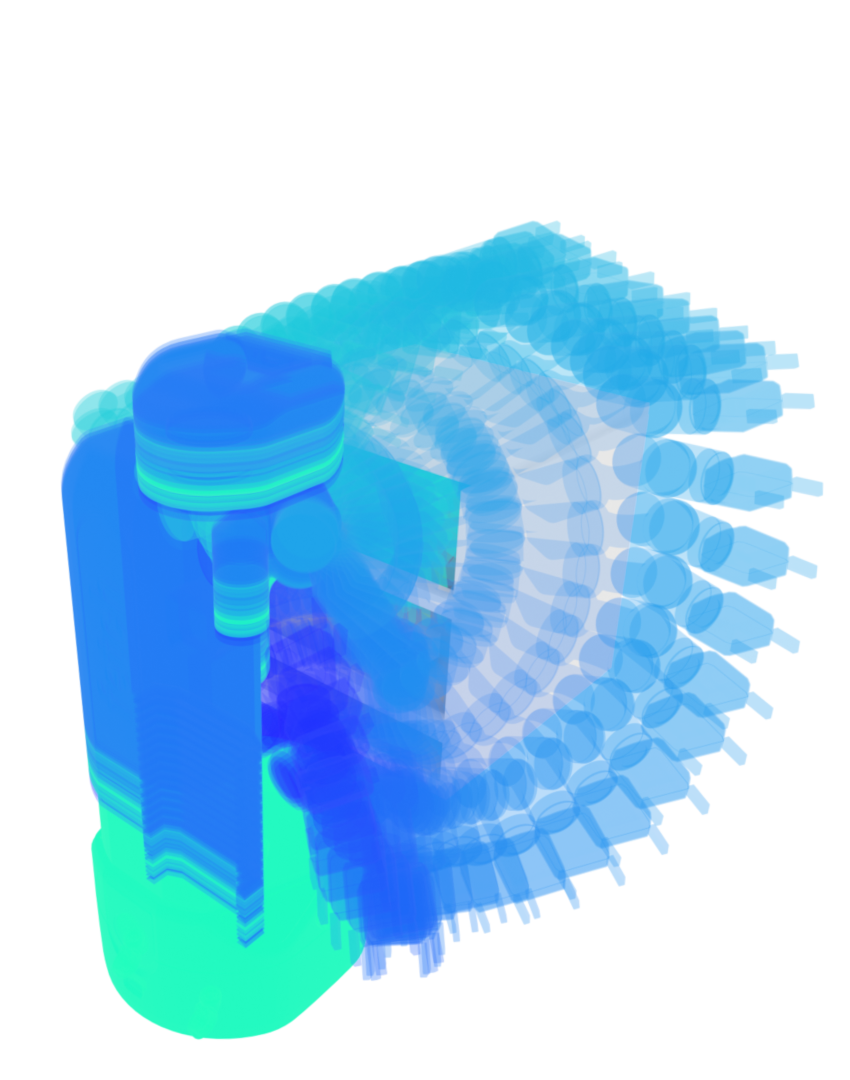}
        \includegraphics[width=.33\linewidth]{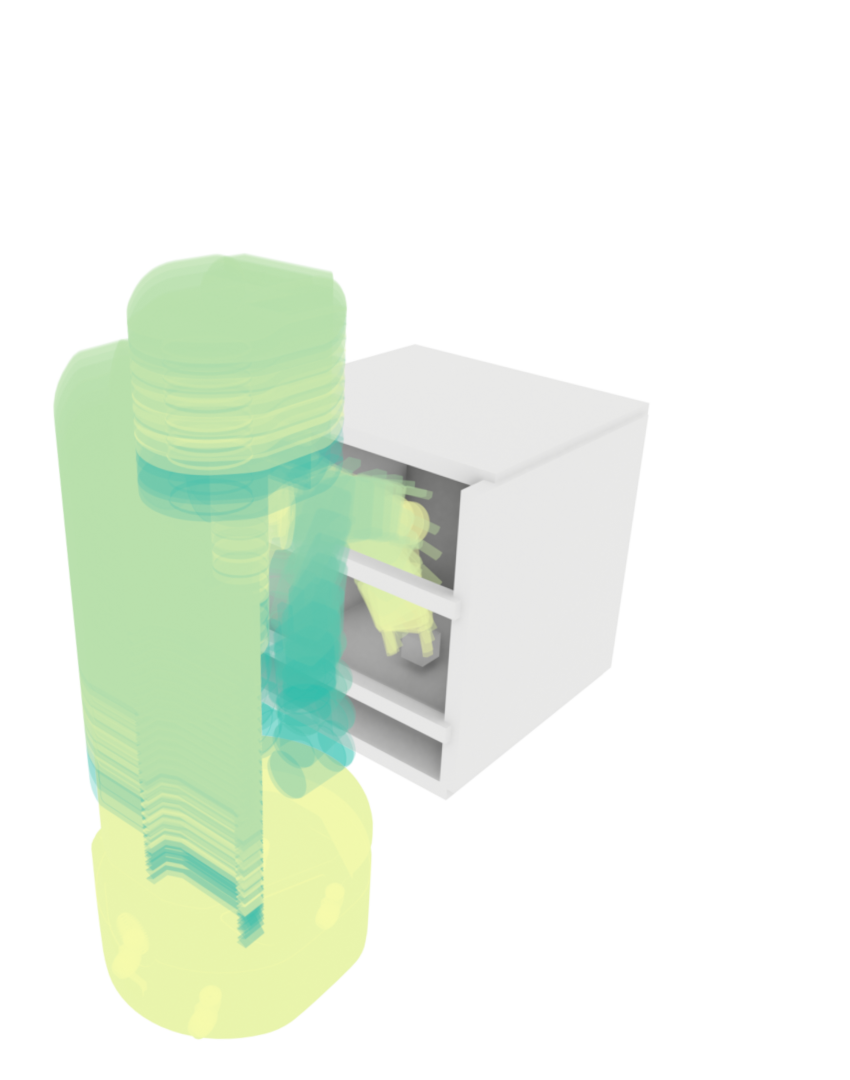}
    }\hfill 

    \subfloat [cage (Panda)] {%
        \includegraphics[width=\linewidth]{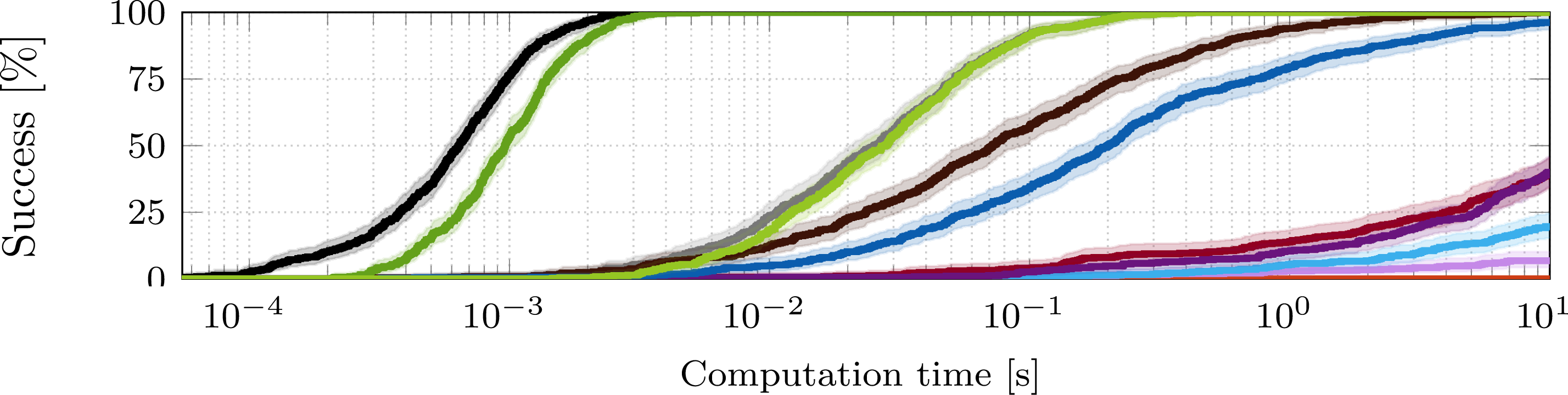}%
    }\hfill
    \subfloat [cage (Fetch)] {%
        \includegraphics[width=\linewidth]{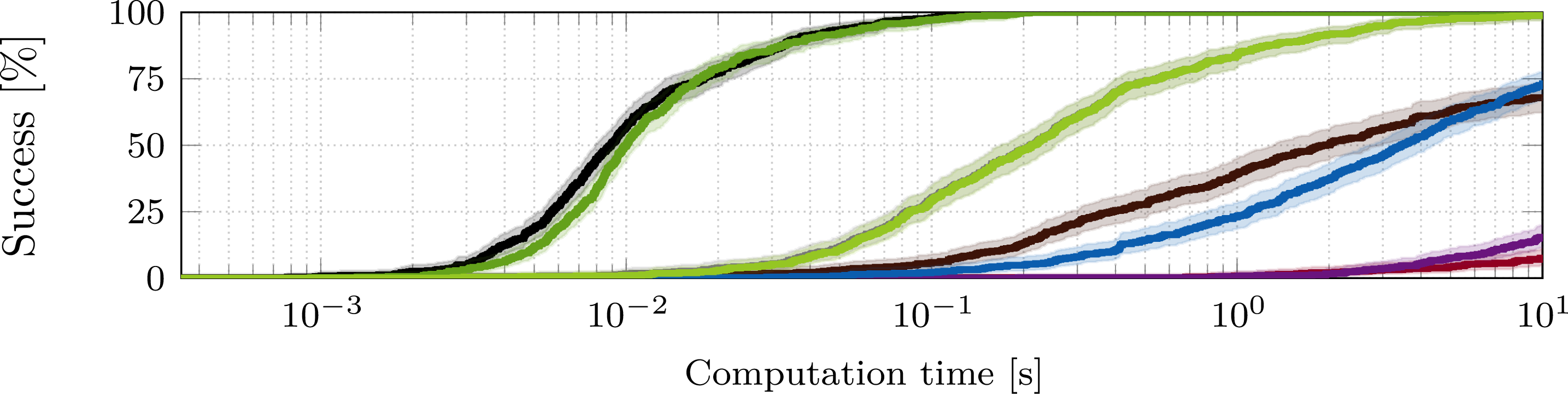}%
    }\hfill
    \subfloat{%
        \includegraphics[width=\linewidth]{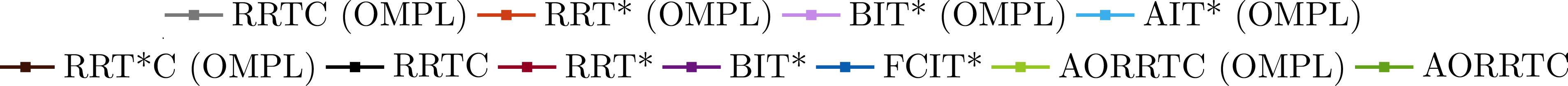}
    }
    
	\caption{
        Results for all planners on the 7 DoF Panda problems, and a subset of planners on the 8 DoF Fetch problems, from the MotionBenchMaker~\cite{mbm} dataset.
        All planners are implemented in either OMPL (labelled with \emph{OMPL}) or are implemented in VAMP.
        The paths for the challenging 8 DoF Fetch \emph{cage} problem (\cref{fig:converge}c) are shown in (a) with the VAMP RRT-Connect solution found after 62ms in orange, the initial VAMP AORRTC solution (\ie RRT-Connect with path simplification) found after 63ms in blue, and the last VAMP AORRTC solution found after 3.5s in green.
        The percentage of trials that found a solution on the \emph{cage} environment within the given time are shown in (b) and (c) with Clopper-Pearson 99\% confidence intervals~\cite{pdt}.
        Both VAMP and OMPL \acs{AORRTC} find initial solutions significantly faster than all other tested VAMP and OMPL \acs{ASAO} planners.
        \squeezeWords
    } \label{fig:cage_initial_all}
    \end{figure} 
    
    Sampling-based planners, such as \ac{PRM}~\cite{prm} and \ac{RRT}~\cite{rrt}, avoid the need for an \emph{a priori} approximation by incrementally sampling the search space.
    This allows them to search an increasingly accurate approximation of the underlying continuous space and has made them a popular and effective choice for high-dimensional problems.
    Sampling-based planners provide probabilistic guarantees.
    Algorithms are said to be probabilistically complete if their probability of finding a solution goes to one as their number of samples approaches infinity, if a solution exists.
    They are said to be \ac{ASAO} if they have probability one of asymptotically converging towards the optimal solution with an infinite number of samples~\cite{prmstar}.

    RRT-Connect~\cite{rrtc} extends \ac{RRT} to interleave searching for a feasible path from both the start and goal.
    This bidirectional satisficing planner is widely used because of its simple implementation and fast initial solution time.
    RRT-Connect provides no solution quality guarantees and does not improve its solution with more samples (\ie it is not \ac{ASAO}).
    Path smoothing or simplification~\cite{shortcut1,shortcut2,bspline, hybrid} can improve the cost of the often low-quality solutions found by RRT-Connect (\cref{fig:cage_initial_all:motions}) but provide no global quality guarantees.

    Anytime \ac{ASAO} planners, such as RRT*~\cite{prmstar} and \ac{BIT*}~\cite{bit}, are probabilistically guaranteed to find a solution and then converge to the optimum.
    These algorithms use additional planning time to improve their current solution, but the overhead required to guarantee almost-sure asymptotic optimality can increase the time required to find an initial solution.
    Research has focused on how to improve initial solution times and the rate of convergence towards the optimum~\cite{ait}.
    \squeezeWords

    The \acl{AOX} meta-algorithm~\cite{aox:1} poses an alternative framing of the optimal planning problem.
    Instead of requiring planners to optimize the cost of a feasible path in configuration space, \acl{AOX} asks planners to find a feasible path through a search space that has been augmented to include an extra dimension. 
    This extra dimension represents the cost to reach each configuration. %
    Calling a satisficing planner on a \emph{sequence} of these augmented search spaces with appropriately decreasing limits in the cost dimension has been proven to be \ac{ASAO}~\cite{aox:1, aox:2}. %
    \squeezeWords

    \ac{AORRTC}\footnote{Pronounced aortic ($\overline{\text{a}}$-\textquotesingle $\dot{\text{o}}$r-tik).} applies the ideas of \acl{AOX} and RRT-Connect to perform a bidirectional search in a cost-augmented search space.
    This finds initial solutions as fast as RRT-Connect and converges towards high-quality solutions orders of magnitudes faster than other \ac{ASAO} algorithms.
    The performance of \ac{AORRTC} is tested with both \ac{OMPL}~\cite{ompl} and \ac{VAMP}~\cite{vamp} implementations. These demonstrate the effectiveness of the approach with and without \ac{SIMD} acceleration and specifically show that \ac{AORRTC} can converge close to the optimum of high-dimensional planning problems in microseconds with \ac{SIMD} acceleration.
	
\begin{figure*}[t]
    
        \includegraphics{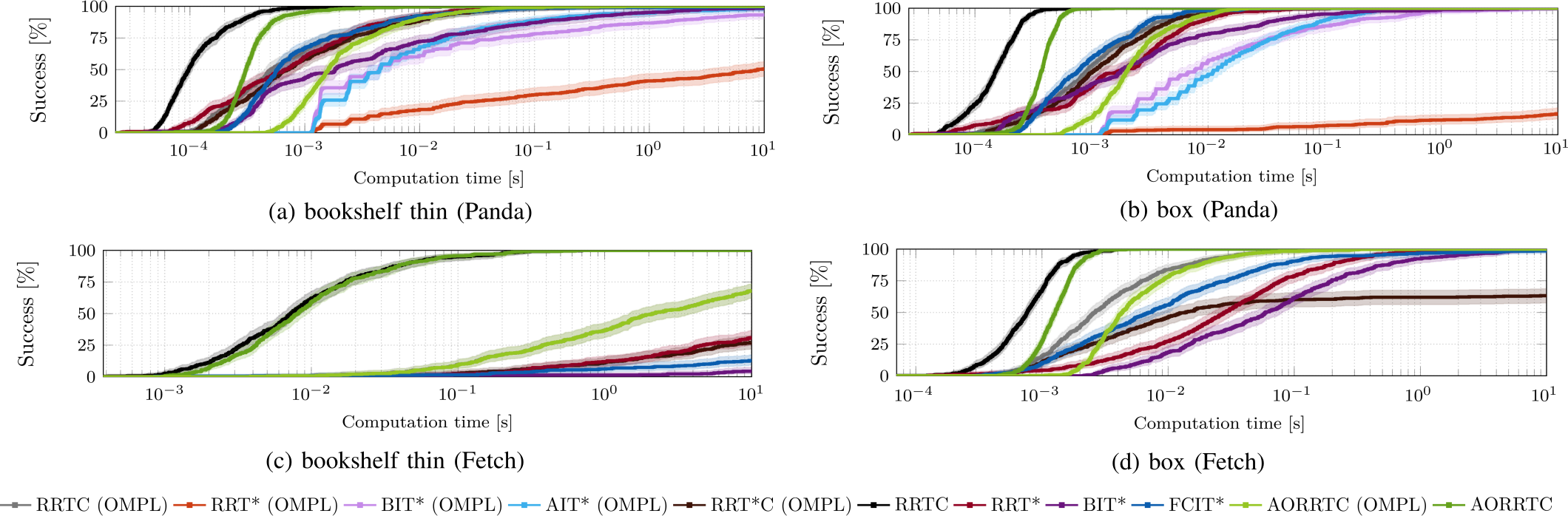}
        \caption{ 
        Initial solution results for 5 trials of all planners on the 7 DoF Panda, and a subset of these planners on the 8 DoF Fetch, for all problems from the \emph{bookshelf thin}, (a) and (c), and \emph{box}, (b) and (d), environments from the MotionBenchMaker~\cite{mbm} dataset (\cref{sec:exp}).
        Each plot shows the percentage of runs that found a solution at any given time with Clopper-Pearson 99\% confidence intervals~\cite{pdt}.
        VAMP \acs{AORRTC} finds initial solutions faster than all other tested \ac{ASAO} planners. 
        \squeezeWords
        }
        \label{fig:quad_initial_all}
    \end{figure*}

    Both implementations of \ac{AORRTC} were evaluated on the 7 DoF Panda and 8 DoF Fetch planning problems from the \ac{MBM} dataset~\cite{mbm}.
    \ac{AORRTC} found initial solutions as fast as RRT-Connect and faster than all tested \ac{ASAO} planners while consistently solving more problems~(\cref{fig:cage_initial_all,fig:quad_initial_all}).
    It also converged to better solutions faster than the tested \ac{ASAO} planners.
    These relative performances held for both OMPL and VAMP implementations of \ac{AORRTC}~(\cref{sec:disc}).

    \section{Related Work} \label{sec:related}
	
    Improving the performance of real-world motion planning often focuses on finding higher-quality solutions in less time.
    There has been significant work improving sampling-based motion planning with these goals in mind.

    \subsection{Almost-Surely Asymptotically Optimal Planning}
	
    Anytime \ac{ASAO} planners use additional planning time to improve their approximation of the search space and find better solutions.
    These planners asymptotically converge to the optimal solution with probability one (\ie almost surely).

    Nearest neighbour lookups and edge evaluations are major computational costs for sampling-based motion planning~\cite{bottleneck}.
    These computationally expensive operations are often performed more frequently in \ac{ASAO} planners to guarantee almost-sure asymptotic optimality which can make them slower to find initial solutions than satisficing planners.
	
    Anytime \ac{ASAO} planners, such as RRT*~\cite{prmstar} and \ac{BIT*}~\cite{bit}, iteratively sample the search space to improve both their approximation and solution. 
    RRT* incrementally rewires new vertices to reduce path costs while \ac{BIT*} searches a batch of samples in an order informed by heuristics to consider states in order of potential solution quality.
    These planners require computational effort to maintain their \ac{ASAO} guarantees and significant work has focused on improving their performance.
    \squeezeWords

    Several RRT* extensions have improved the planner's convergence to an optimal solution.
    RRT\textsuperscript{\#}~\cite{rrtsharp} builds a search tree that contains information about the best cost-to-come to all vertices that could possibly belong to an optimal solution instead of only locally rewiring vertices.
    RRT*-Smart~\cite{rrt*smart} and Informed RRT*~\cite{informed} instead improve the planner's performance by leveraging problem specific information.
    These planners use information from prior search efforts to generate samples more intelligently in order to reduce planning time when improving a solution but this informed sampling cannot help find an initial solution.

    \ac{FCIT*}~\cite{fcit} leverages the reduced cost of edge evaluations enabled by \ac{VAMP}'s \ac{SIMD} parallelism to search a fully connected graph.
	This fully exploits the approximation of the search space and avoids the need to maintain costly nearest neighbour structures.
    FCIT* finds high-quality solutions faster than non \ac{SIMD}-accelerated planners, including RRT-Connect, but cannot find solutions as fast as \ac{SIMD}-accelerated RRT-Connect.

    \ac{AORRTC} is \ac{ASAO} like RRT* and BIT* and improves the quality of its found solution given additional planning time, but finds initial solutions as fast as RRT-Connect.
    Unlike RRT* or RRT\textsuperscript{\#}, \ac{AORRTC} does not have to rewire its tree to maintain its \ac{ASAO} guarantees, instead randomly sampling lower cost bounds when adding a new vertex to potentially connect it to a lower-cost parent.
    \ac{AORRTC} leverages informed sampling as is done in Informed RRT*, but restarts its search after finding a solution rather than continuing to plan with the same tree.
    \squeezeWords

    \subsection{Bidirectional Planning}

    Bidirectional sampling-based planners, such as RRT-Connect~\cite{rrtc}, explore the search space by extending a tree from both the start and goal vertices and trying to connect these trees.
    RRT-Connect finds initial solutions significantly faster than other planning algorithms on most real-world manipulation problems~\cite{compreview} but offers no guarantees on solution quality and often finds low quality solutions.
    \squeezeWords
    
    Some bidirectional \ac{ASAO} planners, such as \ac{B-RRT*}~\cite{brrt*} and RRT*-Connect~\cite{rrt*connect}, extend the fast bidirectional search of RRT-Connect with the \ac{ASAO} guarantees of RRT*.
    These algorithms require RRT*-style rewiring throughout the entire search in order to guarantee almost-surely asymptotic optimality.
    This increase in computational cost increases the time required to find an initial solution relative to RRT-Connect.

    Other bidirectional \ac{ASAO} planners, such as \ac{AIT*} and \ac{EIT*}~\cite{ait}, instead use an asymmetric bidirectional search where information is shared between the reverse and forward searches.
    The lightweight reverse search calculates heuristics from the current samples to inform the computationally expensive forward search, which finds a solution and passes collision checking information to the reverse search to update the heuristics. 
    This reduces edge evaluation costs but its utility depends on the cost of edge evaluations relative to nearest neighbour lookups.
    \squeezeWords

    \ac{AORRTC} performs an \ac{ASAO} bidirectional search similar to \ac{B-RRT*} and RRT*-Connect but does not require RRT*-style rewiring to maintain its \ac{ASAO} guarantees.
    Unlike AIT* and EIT*, \ac{AORRTC} grows both trees and tries to connect them instead of using the reverse search to calculate heuristics.
    \squeezeWords
    
    \subsection{Augmented Search Spaces}

    Augmented search spaces have been used in other planners to improve planning.
    \ac{TB-RRT}~\cite{tbrrt} augments the configuration space with a time dimension for planning with dynamic obstacles or goals.
    This time-augmented space transforms dynamic obstacles in the original problem into static obstacles in the time-augmented search space~\cite{roadmap_dynamic} and simplifies planning in dynamic environments but offers no guarantees for solution time or quality.

    Other planners, such as \ac{WHCA*}~\cite{whca} and \ac{SIPP}~\cite{sipp}, abstract a time-augmented search space to reduce planning time.
    \ac{WHCA*} plans in the time-augmented search space up to a user-specified threshold and then excludes the time dimension and any dynamic obstacles for later times.
    \ac{SIPP} discretizes the continuous time dimension into discrete \emph{safe intervals}, which describe a  duration during which a configuration is considered valid, and plans in this simplified state-\emph{safe interval} augmented search space.
    These search space abstractions help improve planner performance, but offer no guarantees on solution quality. 

    Some bidirectional planners, such as \ac{ST-RRT*}~\cite{strrtstar} and \ac{SI-RRT}~\cite{sirrt}, extend RRT-Connect to search in a time-augmented search space.
    \ac{ST-RRT*} searches an incrementally increasing range in the time dimension to quickly find initial solutions in unbounded time spaces but requires RRT*-style rewiring to maintain its guarantees on solution quality.
    \ac{SI-RRT} instead searches in a simplified state-\emph{safe interval} augmented search space to quickly find initial solutions in high-dimensional dynamic environments but offers no guarantees on solution quality.
    \squeezeWords

    The meta-algorithm \acl{AOX}~\cite{aox:1} extends satisficing kinodynamic planners to include \ac{ASAO} guarantees through the use of a cost-augmented search space.
    This search space consists of the $n$ dimensions of the configuration space and a $\left(n+1\right)^\text{th}$ dimension that describes the cost to reach each state. 
    Satisficing planners can almost-surely converge asymptotically to the optimal solution by finding a series of feasible plans in this augmented search space when the cost function and the dynamics of the robot are Lipschitz continuous~\cite{aox:2}.
    This meta-algorithm has been applied to the \ac{EST}~\cite{aox:1} and \ac{RRT}~\cite{aox:2}.
     
    \ac{AORRTC} searches in an augmented space, like \ac{TB-RRT} and \ac{ST-RRT*}, but this space is augmented with a cost dimension instead of a time dimension.
    Unlike \ac{WHCA*}, \ac{SIPP}, and \ac{SI-RRT}, \ac{AORRTC} does not search an abstraction of its augmented search space.

    \section{\acl{AORRTC}} \label{sec:aorrtc} %

    \ac{AORRTC} applies the ideas of \acl{AOX} and RRT-Connect to create an anytime \ac{ASAO} planner that finds initial solutions as fast as RRT-Connect and asymptotically converges towards the optimal solution with additional planning time in an anytime manner.
    Pseudocode for \ac{AORRTC} is presented in \cref{algo:aox} with changes to RRT-Connect (\crefrange{algo:rrtc}{algo:end}) marked in red.
    \squeezeWords

    \begin{algorithm} [tbp] 
    \caption{AORRTC}\label{algo:aox}
    
    \SetKwProg{algo}{Algorithm}{}{}
    \SetKwProg{func}{Function}{}{}%

    \func{\upshape\texttt{aorrtc}$()$}{
        $\sigma_0 \gets $\upshape\texttt{simplify}$($\upshape\texttt{rrt{-}connect}$(\infty))$\; \label{algo:aox:init}
		$\sigma_\textup{best} \gets \sigma_0$;~
		$c_\textup{min} \gets c(\sigma_0)$\;
		
		\Repeat{\upshape\texttt{stop}}
		{
			$\sigma_i \gets $\upshape\texttt{simplify}$($\upshape\texttt{rrt{-}connect}$(c_\textup{min}))$\; \label{algo:aox:rrtc}
		  \If{$\sigma_i \not \equiv \emptyset$}{
                $\sigma_\textup{best} \gets \sigma_i$\; \label{algo:aox:cmin1}
                $c_\textup{min} \gets c(\sigma_i)$\; \label{algo:aox:cmin2}
            }
		}
		\Return{$\sigma_\textup{best}$}\;
    }{}

      \vspace{0.5em}
      \func{\upshape\texttt{rrt{-}connect}$(c_\textup{max})$}{ \label{algo:rrtc}
		
		$V_a \gets \{(x_\textup{start},\newAlgoLine{0})\}$;~
		$E_a \gets \emptyset$;~
		$T_a \coloneqq \{V_a,E_a\}$\; \label{algo:rrtc:tree_a}
		
		$V_b \gets \{(x_\textup{goal},\newAlgoLine{0})\}$;~
		$E_b \gets \emptyset$;~
		$T_b \coloneqq \{V_b,E_b\}$\; \label{algo:rrtc:tree_b}
		
		\Repeat{\upshape\texttt{timeout}}
		{
			$x_\text{rand} \sim \unif\left(X_\textup{\newAlgoLine{$\hat{f}$}}\right)$\; \label{algo:rrtc:new_config}
            \newAlgoLine{$c_\textup{rand} \sim \unif\left(\left(\hat{g}_{T_a}(x_\text{rand}), c_\textup{max}-\hat{h}_{T_a}(x_\text{rand})\right)\right)$\;} \label{algo:rrtc:new_cost}

            $x_\textup{near} \gets $\upshape\texttt{nearest}$\left(T_a,x_\textup{rand},\newAlgoLine{c_\textup{rand}}\right)$\;
            $x_\textup{new} \gets $\upshape\texttt{steer}$(x_\textup{near},x_\textup{rand})$\;
            
            \If{\upshape\texttt{validate}$(x_\textup{near}, x_\textup{new})$}
		      {
    			$\newAlgoLine{c_\textup{new}} \gets $\upshape\texttt{extend}$\left(T_a, x_\textup{near}, x_\textup{new}\right)$\;
                
                \If{\upshape\texttt{connect}$\left(T_b, x_\textup{new},\newAlgoLine{c_\textup{new}}\right)$}
                {
                    \Return{\upshape\texttt{path}$(T_a, T_b)$}\;
                }
    			
            }
			
			\upshape\texttt{swap}$\left(T_a, T_b\right)$;
		}
		
		\Return{$\emptyset$}\;
        }{}

      \vspace{0.5em}
      \func{\upshape\texttt{extend}$\left(T,x_\textup{near},x_\textup{new}\right)$}{
        \newAlgoLine{
            \Repeat{$x_\textup{near} \equiv  x_\text{p}$ \upshape\textbf{or not} \upshape\texttt{validate}$(x_\textup{near}, x_\textup{new})$} 
            {
                \label{algo:extend:resample_start} 
                $x_\text{p} \gets x_\textup{near}$\; 
                $c_\textup{new} \gets g_T(x_\text{p}) + \hat{c}(x_\text{p}, x_\textup{new})$\;
                $c_\textup{rand} \sim \unif\left((\hat{g}_T(x_\textup{new}),c_\textup{new})\right)$\;
                $x_\textup{near} \gets $\upshape\texttt{nearest}$\left(T,x_\textup{new},c_\textup{rand}\right)$\; 
            } \label{algo:extend:resample_end}
        }
        
        $E \stackrel{+}\gets (x_\textup{p},x_\textup{new})$\;
        $V \stackrel{+}\gets (x_\textup{new},\newAlgoLine{c_\textup{new}})$\;

        \Return{$\newAlgoLine{c_\textup{new}}$}\;
        }{}

    \vspace{0.5em}
    \func{\upshape\texttt{connect}$\left(T_b,x,\newAlgoLine{c}\right)$}{
        \Repeat{$x_\textup{new} \equiv x$ \upshape\textbf{or not} \upshape\texttt{validate}$(x_\textup{near}, x_\textup{new})$}
		{
			$x_\textup{near} \gets $\upshape\texttt{nearest}$\left(T_b,x,\newAlgoLine{c_\textup{max} - c}\right)$\;
            $x_\textup{new} \gets $\upshape\texttt{steer}$(x_\textup{near},x)$\;

            \If{\upshape\texttt{validate}$(x_\textup{near}, x_\textup{new})$}
    		{
    			\upshape\texttt{extend}$\left(T_b, x_\textup{near}, x_\textup{new}\right)$\;
    		}
		}
        \Return{\upshape\texttt{validate}$(x_\textup{near}, x_\textup{new})$}\;
        }{}

      \vspace{0.5em}
        \func{\upshape\texttt{nearest}$\left(T,x,\newAlgoLine{c}\right)$}{
		$x_\textup{near} {\gets}$ \\ 
                
                $\argmin_{v\in V_T}\limits%
    		\left\{%
    			\newAlgoLine{w_x}|| x - x_v ||^2%
    			\newAlgoLine{%
    				+ w_c|c - c_v|^2%
    			}%
    			~\middle|~%
    			\newAlgoLine{%
    				c_v+\hat{c}(x_v,x) < c%
    			}%
    		\right\}$;\hspace{0ex}\\\label{algo:nearest:near}%
		\Return{$x_\textup{near}$}\;
        }{}\label{algo:end}
	\end{algorithm}

    AORRTC runs RRT-Connect iteratively on a series of problems in the augmented search space with an open upper-bound on solution cost.
    This bound is first determined by RRT-Connect's initial solution after path simplification techniques (\eg randomized shortcutting~\cite{shortcut1, shortcut2} and B-spline smoothing~\cite{bspline}) have been applied (\cref{algo:aox:init}).
    AORRTC then uses this bound to pose a new planning problem where the cost-dimension of the augmented space is limited by the cost of the current best solution (\cref{algo:aox:rrtc}).
    AORRTC applies path simplification each time a better solution is found and this bound is lowered accordingly  (\crefrange{algo:aox:rrtc}{algo:aox:cmin2}).
    
    AORRTC grows trees from the start and goal vertices in the augmented search space where each vertex consists of a configuration and its cost-to-come in the respective tree (\cref{algo:rrtc:tree_a,algo:rrtc:tree_b}).
    The current tree is grown towards a random sample from the augmented search space consisting of a random configuration (\cref{algo:rrtc:new_config}) and a randomly sampled cost bound (\cref{algo:rrtc:new_cost}).
    This sampled cost bound is the upper limit on cost for a connection to the sample and limits connections to only those that could contribute to a solution that is higher-quality than the current best solution.
    After a vertex is connected to a tree its sampled cost bound is randomly resampled from a lower range to see if the vertex can be easily connected with lower cost~(\crefrange{algo:extend:resample_start}{algo:extend:resample_end}).
    
    The underlying RRT-Connect search finds incrementally higher-quality solutions by forcing samples to only attempt connections that would satisfy their randomly sampled cost bound (\cref{algo:nearest:near}).
    Once a solution has been found it is simplified and a new upper bound on solution cost is determined.
    The search is then restarted with this tighter cost bound.
    This restricts each subsequent search to paths with higher quality than the current solution and allows the planner to almost-surely asymptotically converge to an optimal solution even if its underlying search is not \ac{ASAO}
    \squeezeWords

    \begin{table*} [tb] 
        \caption{
    		Summary of all planning results on the Panda (7 DoF) and Fetch (8 DoF) robotic arms for the MotionBenchMaker~\cite{mbm} dataset.
    		Each result indicates the percentage of problems solved in the given environment, the median initial solution time across all problems on that environment, and the median initial path length across all problems on that environment.
            The best result for a given robot and framework on each environment is shown in bold.
    	}\label{plan_results:panda}
    	\scriptsize
        \includegraphics[width=\linewidth]{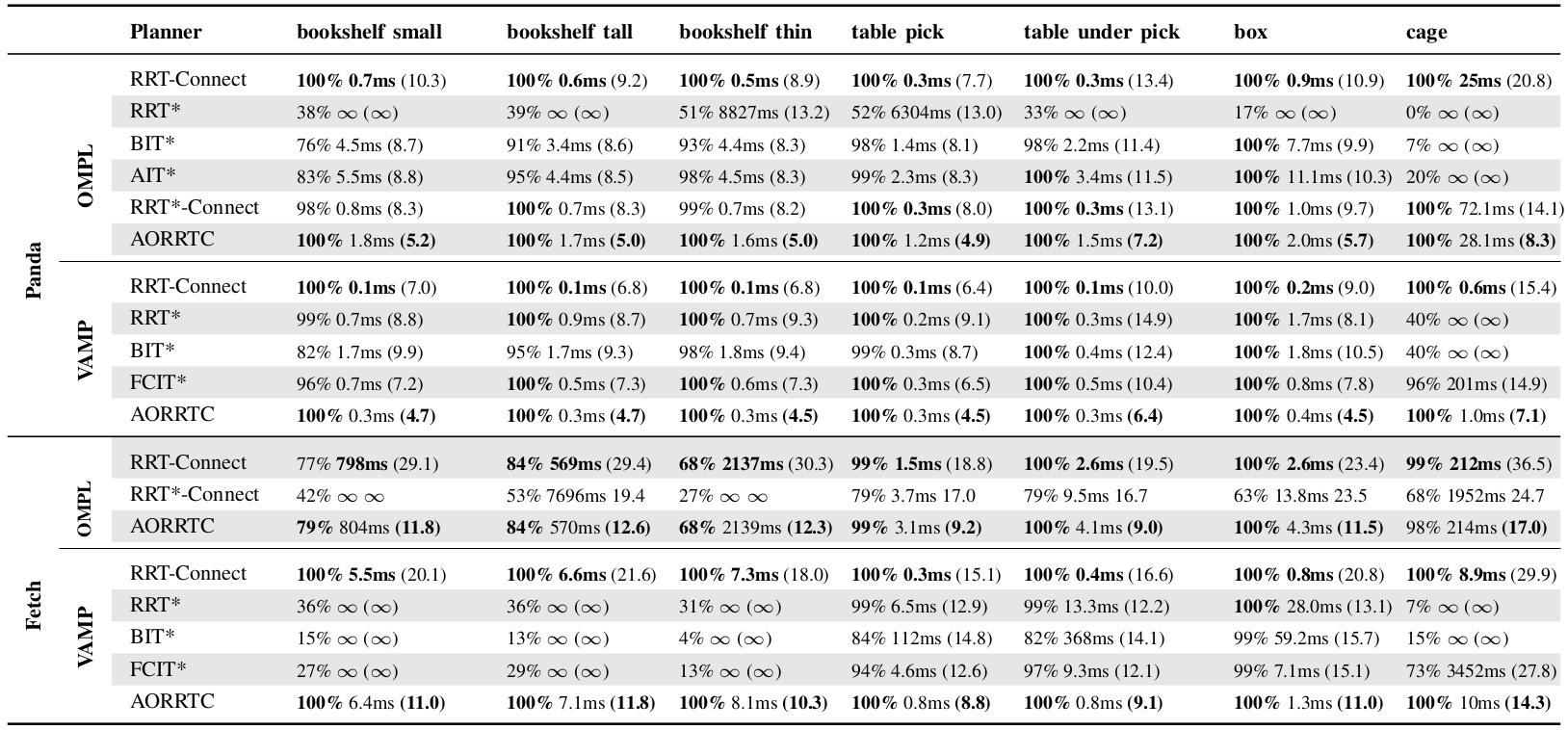}
    \end{table*}

    \subsection{Notation} \label{sec:aorrtc:notation}
	
	We denote the configuration space by $X \subseteq \mathbb{R}^n$ and the invalid and valid subsets as $X_\text{invalid} \subseteq X$ and $X_\text{valid} \coloneqq \mathtt{closure}(X \setminus X_\text{invalid})$, respectively.
	Let $x \in X$ be a configuration, $x_\text{start} \in X_\text{valid}$ be the starting configuration, and $x_\text{goal} \in X_\text{valid}$ be the goal configuration.
	Let $c \in \mathbb{R^+}$ be the cost of an edge or set of consecutive edges (\ie a path) in configuration space.
	We store the searches as trees, $T \coloneqq (V,E)$, each comprising a set of vertices, $V \subseteq \mathbb{R}^{n+1}$, and a set of edges, $E \subseteq V \times V$.
	Each edge, $e \coloneqq (v_\text{p}, v_\text{c}) \in V \times V$, connects two vertices which we refer to as the edge's parent and child, respectively.
	We denote a solution as the sequence of edges, $\sigma = (x_\text{s}, v_0),(v_0,v_1),(v_1,v_2),...,(v_q,x_\text{g})$, where $(v_i,v_j) \in E$.
    \squeezeWords

	The function $c : X \times X \to [0, \infty)$ computes the edge cost between two configurations or, with a slight abuse of notation, the cost of a given solution.
	The function \mbox{$\hat{c}: X \times X \to [0, \infty)$} is an admissible estimate of the edge cost, \ie $\forall x_\text{p}, x_\text{c} \in X, \; \hat{c}(x_\text{p}, x_\text{c}) \leq c(x_\text{p}, x_\text{c})$.
	The function $g_T: V_T \to [0, \infty)$ represents the cost-to-come through a tree, $T$, from the root to the given vertex, $v$.
	The function $\hat{g}_T: X \to [0, \infty)$ represents an admissible estimate of this cost-to-come, \ie $\forall x \in X, \; \hat{g}_T(x) \leq g_T(x)$.
    The function $\hat{h}_{T}: X \to [0, \infty)$ represents the estimated cost-to-go from the configuration, $x$, to the goal of the tree, $T$.
    Note that the goal of one tree, $T_a$, is the root of the other, $T_b$, \ie $\forall x \in X, \hat{h}_{T_a}(x) \equiv \hat{g}_{T_b}(x)$.
    Let $X_{\hat{f}} \subseteq X_\text{free}$ be the informed set of free configurations, \ie $\forall x \in X_{\hat{f}}, g_{T}(x) + \hat{h}_{T}(x) < c_\text{max}$.

    Let $w_x$ and $w_c$ be the distance function weighting for configuration and cost distance, respectively.
	Let $A$ and $B$ be two sets. 
	The notation $\unif(A)$ is shorthand for drawing a sample uniformly from a set, $A$.
	The notation $A \stackrel{+}\gets B$ is shorthand for the set compounding operation, $A \gets A \cup B$.

    \subsection{Augmented Search Space} \label{sec:aorrtc:augment}
	\ac{AORRTC} searches an augmented $(n+1)$-dimensional search space where the first $n$ dimensions are the configuration space and the $(n+1)^\text{th}$ dimension is the cost to reach that configuration.
	New configurations are sampled uniformly from configuration space (\cref{algo:rrtc:new_config}).
    The upper bound on the cost of a new state, $c_\textup{rand}$, is randomly sampled uniformly between the minimum and maximum possible cost of a solution passing through the state such that the state is feasible and useful, \ie $\hat{g}(x) + \hat{h}(x) \leq c_\textup{rand} < c_\text{max}$ 
    (\cref{algo:rrtc:new_cost}).
    This cost is an upper bound for a connection to the configuration to be considered valid in the augmented space and the true cost of a configuration will be calculated from its parent configuration.
    \squeezeWords
    
    The nearest neighbour in the augmented search space is defined as the state closest in both configuration and cost (\cref{algo:nearest:near})~\cite{aox:1, aox:2}.
    This distance function means samples with high cost bounds are closest to vertices with high costs-to-come, and can inflate the cost of reaching the new configuration.
    To address this, the cost of a newly added vertex is resampled to search for lower cost neighbours after it is added to the tree (\crefrange{algo:extend:resample_start}{algo:extend:resample_end}).
    This resampling continues until an invalid edge or the same parent is found.
    The cost of a new vertex, $v$, in the tree is the cost-to-come through its parent vertex, $v_\text{p}$, \ie $g_T(v) = g_T(v_\text{p}) + c(v_\text{p},v)$.

    \rowcolors{1}{white}{white}
    \begin{figure*}[t]
        \includegraphics{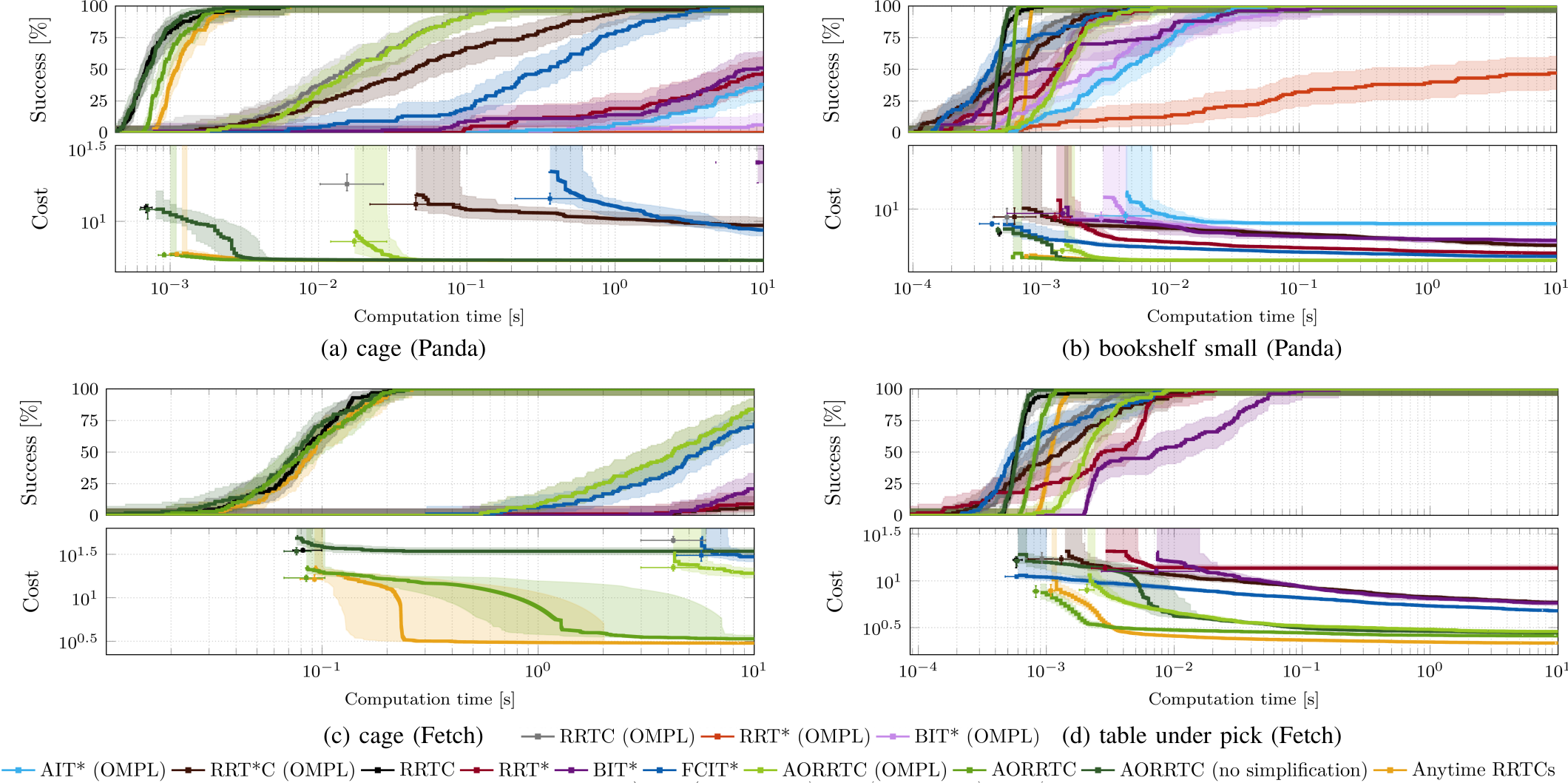}
    	\caption{ 
            Convergence results for 100 trials of all planners on the 7 DoF Panda, and a subset of these planners on the 8 DoF Fetch, for a single problem from the \emph{cage}, (a) and (c), \emph{bookshelf small}, (b), and \emph{table under pick}, (d), environments from the MotionBenchMaker~\cite{mbm} dataset (\cref{sec:exp}).
    		The top plot of each subfigure shows the percentage of runs that found a solution by a given time with Clopper-Pearson 99\% confidence intervals,
    		and the bottom shows the median initial path length and median path length over time with nonparametric 99\% confidence intervals~\cite{pdt}.
        }
    	\label{fig:converge}
    \end{figure*}

    \subsection{Analysis} \label{sec:aorrtc:analysis}

    The \acl{AOX} meta-algorithm has been proven by \cite{aox:2} to be \ac{ASAO} when the cost function and robot dynamics are Lipschitz continuous and the planner used on the augmented search space is probabilistically complete.
    AORRTC is therefore \ac{ASAO} if the version of RRT-Connect it uses is probabilistically complete.
    The original RRT-Connect is probabilistically complete~\cite{rrtc} and the added cost resampling (\crefrange{algo:extend:resample_start}{algo:extend:resample_end}) in AORRTC does not change the vertices in the trees but only the cost of their connections.
    This maintains the probabilistic completeness of RRT-Connect and AORRTC is therefore probabilistically complete and \ac{ASAO} with respect to the original planning problem.

    \subsection{Anytime RRT-Connects} \label{sec:aorrtc:anytime}

    A na\"{i}ve approach to making RRT-Connect an anytime \ac{ASAO} planner would be to run it iteratively with informed sampling to try and find higher-quality solutions.
    This bidirectional extension of Anytime RRTs~\cite{anytimerrts} converges to a reasonable solution on many problems but is not sufficient to probabilistically converge to optimal solutions and is provably \emph{not} \ac{ASAO}~\cite{prmstar}.
    The performance of this \emph{Anytime RRT-Connects} is presented in \cref{fig:converge}, even though it provides no formal optimality guarantees despite its practical performance.
    \squeezeLine

    \subsection{Implementation} \label{sec:aorrtc:implementation}

    \cref{algo:aox} describes a conceptual version of AORRTC that leaves room for several improvements.
    \ac{AORRTC} should sample configurations using direct informed sampling~\cite{informed} when available or rejection sampling otherwise, and should use a balanced bidirectional search~\cite{balanced}.
    Nearest neighbour structures and lookups should be used where appropriate.
    \squeezeWords

     \begin{figure}
    	\centering
        \subfloat [cage (Panda)] {%
    		\includegraphics[width=\linewidth]{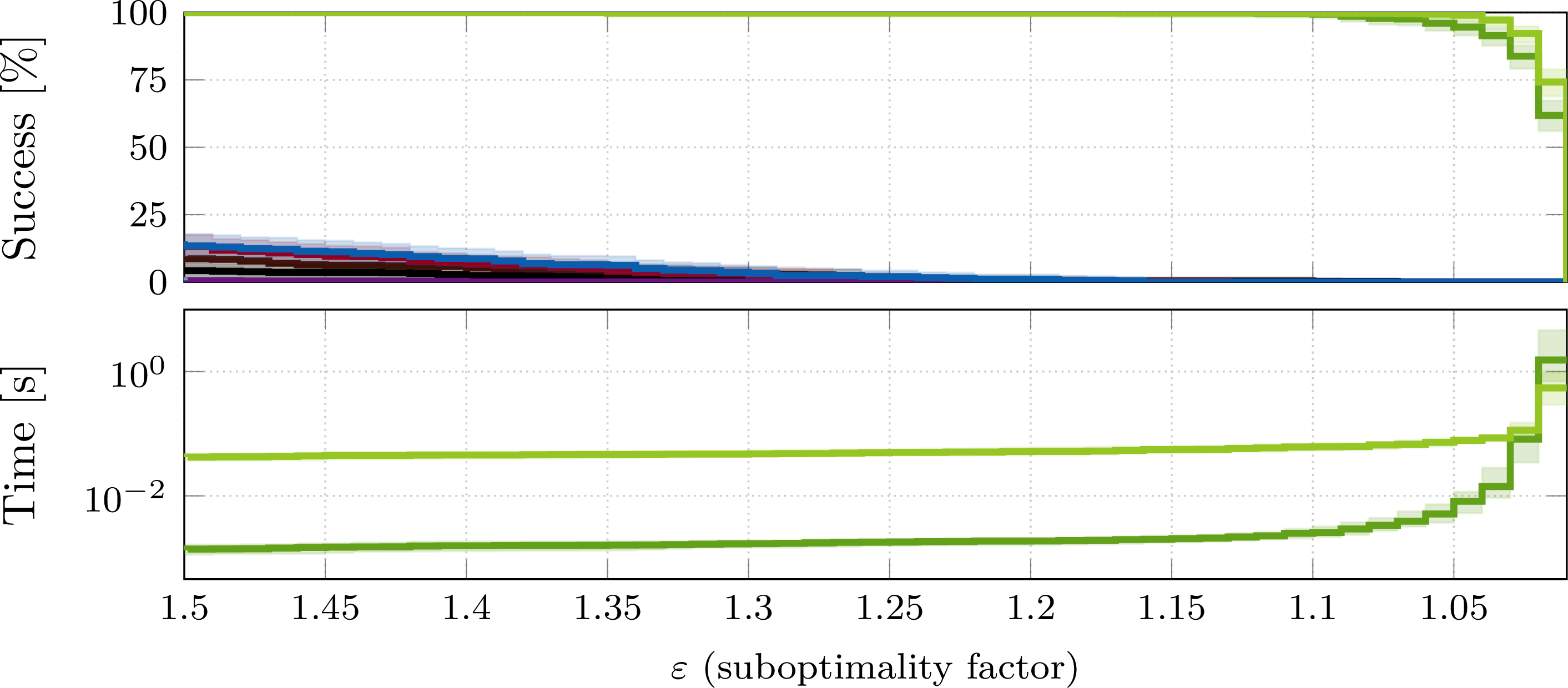}%
    	}\hfill
        \subfloat [bookshelf small (Panda)] {%
    		\includegraphics[width=\linewidth]{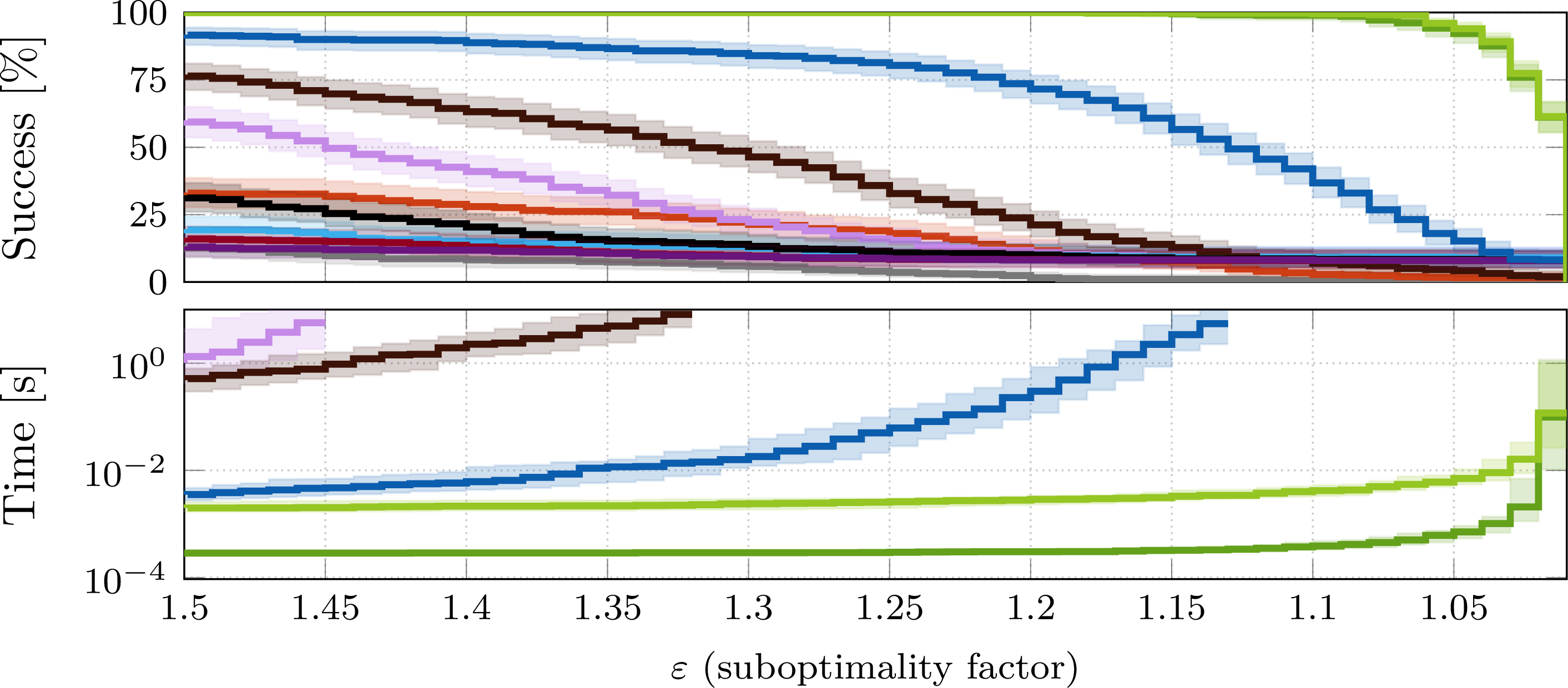}%
    	}\hfill
        \subfloat [table pick (Panda)] {%
    		\includegraphics[width=\linewidth]{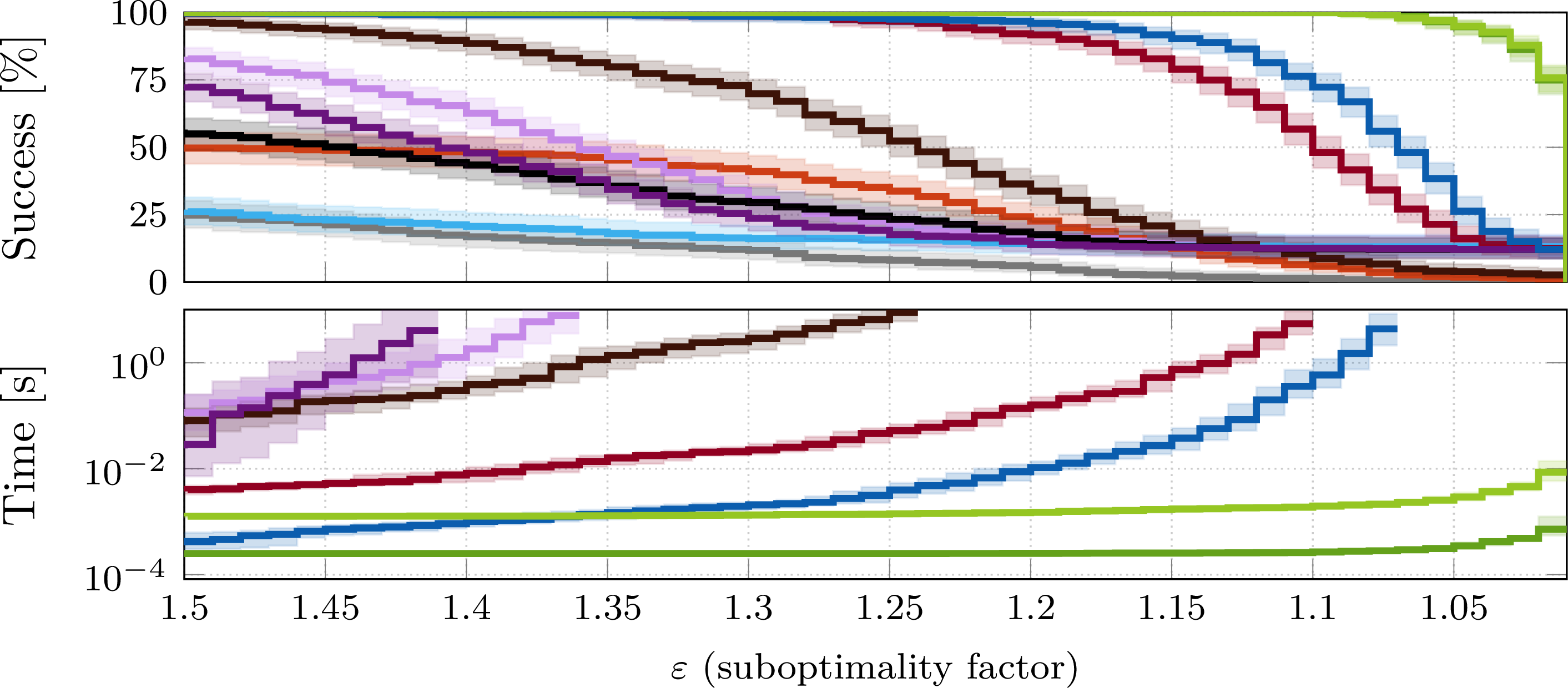}%
    	}\hfill
    	\subfloat{%
    		\includegraphics[width=\linewidth]{images/label_initial_stacked.png}%
    	}
    	
    	\caption{
                Near-optimality results for 5 trials of all planners on the 7 DoF Panda for all problems from the \emph{cage}, (a), \emph{bookshelf small}, (b), and \emph{table pick}, (c), environments from the MotionBenchMaker~\cite{mbm} dataset (\cref{sec:exp}).
                The top plots show the percentage of runs that found a solution within a suboptimality factor, $\epsilon$, of the empirically determined  optimum, $\widehat{c^*}$, with Clopper-Pearson 99\% confidence intervals.
                The bottom plots show the median time to find a solution within a suboptimality factor, $\epsilon$, of the empirically determined optimum, $\widehat{c^*}$, with nonparametric 99\% confidence intervals.
            \squeezeWords
    	} \label{fig:scene_converge}
    \end{figure}

    \section{Experiments} \label{sec:exp}

	OMPL and VAMP implementations of \ac{AORRTC} were evaluated against \ac{OMPL}~\cite{ompl} implementations of RRT-Connect, RRT*, BIT*, AIT*, and RRT*-Connect and VAMP~\cite{vamp} implementations of RRT-Connect, RRT*, BIT*, and FCIT*.
    We also show convergence results for a VAMP implementation of Anytime RRT-Connects (\cref{sec:aorrtc:implementation}) and a variation of VAMP AORRTC without simplification in \cref{fig:converge}.
	The reported solution costs and times for \ac{AORRTC} include the computational costs of randomized shortcutting~\cite{shortcut1, shortcut2} and B-spline smoothing~\cite{bspline}.
    The reported  solution costs and times for the implementations of RRT-Connect do not include simplification but these can be inferred from the initial solution time and cost of AORRTC since this is a single simplified RRT-Connect search.
    Both RRT-Connect and AORRTC used a balanced bidirectional search~\cite{balanced}.
    The implementations of AORRTC and RRT-Connect uses an edge length of 2 (VAMP) and 2.6 (OMPL).
    All planners calculate their RGG constant with Theorem 38 from \cite{prmstar} when applicable.
    A value of 1.0 was used for both distance function weights, $w_x$ and $w_c$.
    \squeezeWords
    
	The planners were tested on the \ac{MBM}~\cite{mbm} dataset, which contains 7 planning environments for different robots where each environment has 100 different planning problems\footnote{Some of the problems are invalid for the Panda and Fetch robots resulting in 699 and 679 total problems, respectively.}.
	These environments cover a range of manipulation planning scenarios, including reaching (\textit{bookshelf tall}, \textit{bookshelf small}, and \textit{bookshelf thin}), constrained reaching (\textit{box} and \textit{cage}), and tabletop manipulation (\textit{table pick} and \textit{table under pick}).
    
	The OMPL and VAMP planners were run on the 7 \ac{DoF} Panda robotic arm \ac{MBM} problems while only the VAMP planners and OMPL versions of AORRTC, RRT*-Connect, and RRT-Connect were run on the harder 8 \ac{DoF} Fetch Mobile Manipulator \ac{MBM} problems.
    Most \ac{OMPL} planners were not tested on the harder Fetch problems because they were not able to reliably find solutions in the available planning time.
    All tests were run in Ubuntu 22.04 on a Intel i7-9750H CPU with 32GB of RAM, and the planning algorithms are implemented in C++17.
	All planners use the default VAMP and \ac{OMPL} samplers with different seeding for each trial.
    The \ac{OMPL} planners use VAMP's collision checking backend.
	The planners were given 10 seconds to evaluate each problem.
    The time and cost axes for all figures are in logarithmic scale.
    \squeezeWords
    
    The success rate for all problems in a given environment are shown in \cref{fig:cage_initial_all,fig:quad_initial_all}.
    \cref{fig:converge} shows the success rate and median solution cost over time for 100 trials of each planner on a single problem for the Panda and Fetch robotic arms.
    \squeezeWords
    
    \cref{fig:scene_converge} shows the success rate and median time to converge to a \emph{near-optimal} solution for 5 trials of each planner on 100 problems from the \emph{cage}, \emph{bookshelf small}, and \emph{table pick} environments for the Panda robotic arm.
    These results compare the number of problems where a solution was found that falls within a suboptimality factor, $\varepsilon$, of an empirical estimate of the optimum cost, $\widehat{c^*}$, as well as the median time to find a solution that satisfies that bound.
    The empirical optimum for a given problem was taken as the minimum cost found across more than 700 trials during development and experiments.
    \squeezeWords

    \section{Discussion} \label{sec:disc}

    OMPL and VAMP \ac{AORRTC} were the only tested \ac{ASAO} planners that found solutions to all Panda problems.
    VAMP \ac{AORRTC} was also the only tested \ac{ASAO} planner that found solutions to all Fetch problems.
    The other tested \ac{ASAO} planners struggled to find solutions on the difficult Fetch problems.
    VAMP RRT-Connect also found solutions to all tested problems, but is not \ac{ASAO}
    Although OMPL \ac{AORRTC} was not able to find solutions to all the Fetch problems in the allowed time, it found solutions to significantly more Fetch problems and found initial solutions to these problems in less time than all tested planners other than RRT-Connect and VAMP \ac{AORRTC}.
    \squeezeWords
    
	The \ac{VAMP} implementation of \ac{AORRTC} outperforms all other tested \ac{ASAO} planners and finds initial solutions in significantly less time on all environments (\cref{plan_results:panda}).
    None of the tested \ac{ASAO} planners converge to higher quality solutions than those found by \ac{VAMP} \ac{AORRTC} (\cref{fig:converge}).
    \ac{OMPL} \ac{AORRTC} found initial solutions faster than most tested OMPL a.s.a.o. planners.
     Some \ac{VAMP} a.s.a.o. planners, including VAMP AORRTC, were able to find solutions faster than OMPL AORRTC (\cref{fig:quad_initial_all}).
    This is because of the performance improvements of VAMP's \ac{SIMD}-accelerated edge evaluation.
    
    The only OMPL a.s.a.o. planner with similar initial solution performance to OMPL AORRTC was OMPL RRT*-Connect.
    The VAMP and OMPL implementations of RRT-Connect also demonstrated similar initial solution performance to the VAMP and OMPL implementations of AORRTC, respectively.
    RRT-Connect and RRT*-Connect found initial solutions microseconds faster than \ac{AORRTC}. %
    This is expected because the RRT-Connect and RRT*-Connect results do not include the computational cost of path simplification while the initial result found by AORRTC is equivalent to a simplified solution found by RRT-Connect.
    This is supported by the variation of AORRTC with no simplification, which finds initial solutions as fast as RRT-Connect and of similar quality (\cref{fig:converge}). 
    RRT-Connect cannot improve its initial solution given additional planning time.
    OMPL RRT*-Connect converged slower and to significantly lower-quality solutions than OMPL AORRTC (\cref{fig:converge}).
    \squeezeWords

    Both OMPL and VAMP \ac{AORRTC} converged to better solutions in significantly less time than all other tested planners (\cref{fig:scene_converge}).
    AORRTC reliably converged to solutions that were close to the empirical optimum within milliseconds, even on the difficult \textit{cage} problems, and converged to these solutions in less time than any other planner.
    The variation of VAMP AORRTC with no simplification also converged to better solutions in significantly less time than all other tested a.s.a.o. planners other than VAMP AORRTC on almost all problems (\cref{fig:converge}).
    This demonstrates that AORRTC's convergence does not rely on simplification, but is accelerated by it.
    \squeezeWords

    \section{Conclusion} \label{sec:conc}

    \ac{AORRTC} uses the ideas of the \acl{AOX} meta-algorithm~\cite{aox:1,aox:2} and RRT-Connect~\cite{rrtc} to design a bidirectional anytime \ac{ASAO} algorithm that quickly finds initial solutions and then converges towards the optimal solution.
	
    \ac{AORRTC} searches an $(n+1)$-dimensional augmented search space, where the first $n$ dimensions are the configuration space and the $(n+1)^\text{th}$ dimension is its cost-to-come.
    It calls a satisficing planner on a sequence of these augmented search spaces with the cost dimension bounded by the current best solution cost to iteratively find higher-quality solutions.
    This has been proven to be \ac{ASAO}~\cite{aox:1,aox:2}.

    \ac{AORRTC} finds initial solutions on the same order of magnitude as RRT-Connect, which is faster than almost all other \ac{ASAO} planners.
    It then uses the remaining planning time to find higher quality solutions than all other tested \ac{ASAO} algorithms in an anytime manner.
    This is demonstrated with and without \ac{SIMD}-acceleration on hundreds of problems across seven different planning environments for both the 7 DoF Panda and 8 DoF Fetch robotic arms.
    \squeezeWords

    AORRTC represents a different approach to \ac{ASAO} planning.
    The majority of previous \ac{ASAO} algorithms approximated the continuously valued search space with increasing accuracy in order to find better solutions with additional planning time and almost-surely asymptotically converge towards the optimum.
    AORRTC instead quickly samples from the set of feasible solutions that could provide a better solution than the current one.
    This avoids the computational cost of maintaining high-resolution approximations of the search space and allows AORRTC to find initial solutions quickly and then rapidly converge towards the optimum.
    AORRTC shows the promise of this alternative approach to \ac{ASAO} planning and we expect that future work will explore its full implications.
    \squeezeWords

    Future work will explore the performance of different distance functions and apply AORRTC to optimize cost functions other than path length, including multivariate and non-smooth cost functions, as well as extending AORRTC to plan in non-Euclidean spaces, or in real-world dynamic environments.
    Information on the implementation of AORRTC is available at \url{https://robotic-esp.com/code/aorrtc/}.
    \squeezeWords

    \printbibliography{}

\end{document}